\documentclass[11pt,letterpaper]{article}
\usepackage{emnlp2016}
\usepackage{times}
\usepackage{latexsym}
\usepackage{graphicx}
\usepackage{amsmath}
\usepackage{amssymb}

\emnlpfinalcopy

\title{Joint Representation Learning of Text and Knowledge for Knowledge Graph Completion}

\author{Xu Han$^{1}$, Zhiyuan Liu$^{2}$, Maosong Sun$^{2,3}$ \\
$^{1}$ Department of Computer Science and Technology, Tsinghua University, Beijing, China\\
$^{2}$ National Lab for Information Science and Technology, \\
State Key Lab on Intelligent Technology and Systems, \\
Department of Computer Science and Technology, Tsinghua University, Beijing, China \\
$^{3}$ Jiangsu Collaborative Innovation Center for Language Ability, \\
Jiangsu Normal University, Xuzhou 221009 China \\
}

\date{}

\begin{document}

\maketitle

\begin{abstract}
  Joint representation learning of text and knowledge within a unified semantic space enables us to perform knowledge graph completion more accurately. In this work, we propose a novel framework to embed words, entities and relations into the same continuous vector space. In this model, both entity and relation embeddings are learned by taking knowledge graph and plain text into consideration. In experiments, we evaluate the joint learning model on three tasks including entity prediction, relation prediction and relation classification from text. The experiment results show that our model can significantly and consistently improve the performance on the three tasks as compared with other baselines.
\end{abstract}

\section{Introduction}
\label{intro}

People construct various large-scale knowledge graphs (KGs) to organize structural knowledge about the world, such as Freebase \cite{bollacker2008freebase}, YAGO \cite{suchanek2007yago}, DBPedia \cite{auer2007dbpedia} and WordNet \cite{miller1995wordnet}. A typical knowledge graph is a multiple-relational directed graph with nodes corresponding to entities, and edges corresponding to relations between these entities. Knowledge graphs are playing an important role in numerous applications such as question answering and Web search.

The facts in knowledge graphs are usually recorded as a set of relational triples ($h$, $r$, $t$) with $h$ and $t$ indicating \emph{head} and \emph{tail} entities and $r$ the relation between $h$ and $t$, e.g., (\emph{Mark Twain}, \texttt{PlaceOfBirth}, \emph{Florida}).

Typical large-scale knowledge graphs are usually far from complete. The task of knowledge graph completion aims to enrich KGs with novel facts. Based on the network structure of KGs, many graph-based methods have been proposed to find novel facts between entities \cite{lao2011random,lao2010relational}. Many efforts are also devoted to extract relational facts from plain text \cite{zeng2014relation,dos2015classifying}. However, these approaches cannot jointly take both KGs and plain texts into consideration.

\begin{figure}[]
\centering
\includegraphics[width=\columnwidth]{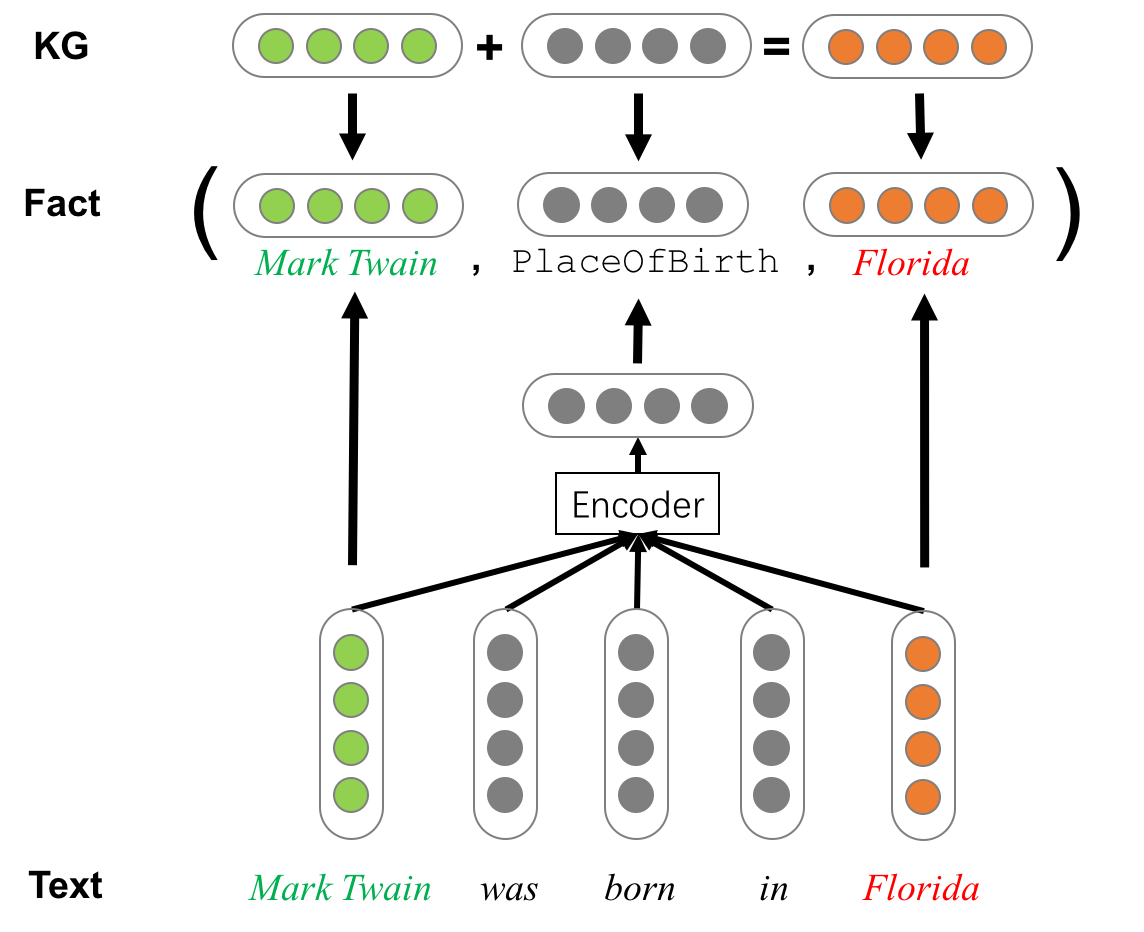}
\caption{The framework of joint representation learning of text and knowledge.}
\label{fig:joinglearning}
\end{figure}

In recent years, neural-based knowledge representation has been proposed to encode both entities and relations into a low-dimensional space, which are capable to find novel facts \cite{bordes2013translating,wang2014transh,lin2015learning}. More importantly, neural models enable us to conduct joint representation learning of text and knowledge within a unified semantic space, and perform knowledge graph completion more accurately.

Some pioneering works have been done. For example, \cite{wang2014knowledge} performs joint learning simply considering alignment between words and entities, and \cite{toutanova2015representing} extracts textual relations from plain texts using dependency parsing to enhance relation embeddings. These works either consider only partial information in plain text (entity mentions in \cite{wang2014knowledge} and textual relations in \cite{toutanova2015representing}), or rely on complicated linguistic analysis (dependency parsing in \cite{toutanova2015representing}) which may bring inevitable parsing errors.

To address these issues, we propose a novel framework for joint representation learning. As shown in Figure \ref{fig:joinglearning}, the framework is expected to take full advantages of both text and KGs via complicated alignments with respect to words, entities and relations. Moreover, our method applies deep neural networks instead of linguistic analysis to encode the semantics of sentences, which is especially capable of modeling large-scale and noisy Web text.

We conduct experiments on a real-world dataset with KG extracted from Freebase and text derived from the New York Times corpus. We evaluate our method on the tasks including entity prediction, relation prediction with embeddings and relation classification from text. Experiment results demonstrate that, our method can effectively perform joint representation learning and obtain more informative knowledge representation, which significantly outperforms other baseline methods on all three tasks.

\section{Related Work}
\label{sec:related}
The work in this paper relates to representation learning of KGs, words and textual relations. Related works are reviewed as follows.

\textbf{Representation Learning of KGs.} A variety of approaches have been proposed to encode both entities and relations into a continuous low-dimensional space. Inspired by \cite{mikolov2013distributed}, TransE \cite{bordes2013translating} regards the relation $r$ in each ($h$, $r$, $t$) as a translation from $h$ to $t$ within the low-dimensional space, i.e., $\textbf{h} + \textbf{r} = \textbf{t}$, where $\textbf{h}$ and $\textbf{t}$ are entity embeddings and $\textbf{r}$ is relation embedding. Despite of its simplicity, TransE achieves the state-of-the-art performance of representation learning for KGs, especially for those large-scale and sparse KGs. Hence, we simply incorporate TransE in our method to handle representation learning for KGs.

Note that, our method is also flexible to incorporate extension models of TransE, such as TransH \cite{wang2014transh} and TransR \cite{lin2015learning}, which is not the focus of this paper and will be left as our future work.

\textbf{Representation Learning of Textual Relations.} Many works aim to extract relational facts from large-scale text corpora \cite{mintz2009distant,riedel2010modeling}. This indicates textual relations between entities are contained in plain text. In recent years, deep neural models such as convolutional neural networks (CNN) have been proposed to encode semantics of sentences to identify relations between entities \cite{zeng2014relation,dos2015classifying}. As compared to conventional models, neural models are capable to accurately capture textual relations between entities from text sequences without explicitly linguistic analysis, and further encode into continuous vector space. Hence, in this work we apply CNN to embed textual relations and conduct joint learning of text and KGs with respect to relations.

Many neural models such as recurrent neural networks (RNN) \cite{zhang2015relation} and long-short term memory networks (LSTM) \cite{xu2015classifying} have also been explored for relation extraction. These models can also be applied to perform representation learning for textual relations, which will be explored in future work.

\textbf{Representation Learning of Words.} Given a text corpus, we can learn word representations without supervision. The learning objective is defined as the likelihood of predicting its context words of each word or vice versa \cite{mikolov2013distributed}. Continuous bag-of-words (CBOW) \cite{mikolov2013efficient} and Skip-Gram \cite{mikolov2013linguistic} are state-of-the-art methods for word representation learning. The learned word embeddings can capture both syntactic and semantic features of words derived from plain text. As reported in many previous works, deep neural network will benefit significantly if being initialized with pre-trained word embeddings \cite{erhan2010does}. In this work, we apply Skip-Gram for word representation learning, which serves as initialization for joint representation learning of text and KGs.

\section{The Framework}
In this section we introduce the framework of joint representation learning, starting by notations and definitions.

\subsection{Notations and Definitions}

We denote a knowledge graph as $G = \{E, R, T\}$, where $E$ indicates a set of entities, $R$ indicates a set of relation types, and $T$ indicates a set of fact triples. Each triple $(h, r, t) \in T$ indicates there is a relation $r \in R$ between $h \in E$ and $t \in E$.

We denote a text corpus as $D$ and its vocabulary as $V$, containing all words, phrases and entity mentions. In the corpus $D$, each sentence is denoted as a word sequence $s = \{x_1, \ldots, x_n\}, x_i \in V$, and the length is $n$.

For entities, relations and words, we use the bold face to indicate their corresponding low-dimensional vectors. For example, the embeddings of $h, t \in E$, $r \in R$ and $x \in V$ are $\mathbf{h}, \mathbf{t}, \mathbf{r}, \mathbf{x} \in \mathbb{R}^{k}$ of $k$ dimension, respectively.

\subsection{Joint Learning Method}
As mentioned in Section \ref{sec:related}, representation learning methods have been proposed for knowledge graphs and text corpora respectively. In this work, we propose a joint learning framework for both KGs and text.

In this framework, we aim to learn representations of entities, relations and words jointly. Denote all these representations as model parameters $\theta = \{\theta_E, \theta_R, \theta_V\}$. The framework aims to find optimized parameters
\begin{equation}
\hat{\theta} = {\arg\min}_{\theta} \mathcal{L}_{\theta}(G, D),
\end{equation}
where $\mathcal{L}_{\theta}(G, D)$ is the loss function defined over the knowledge graph $G$ and the text corpus $D$. The loss function can be further decomposed as follows,
\begin{equation}
\mathcal{L}_{\theta}(G, D) = \mathcal{L}_{\theta_E, \theta_R}(G) + \tau \mathcal{L}_{\theta}(D) + \lambda \lVert \theta \rVert_2,
\end{equation}
where $\tau$ and $\lambda$ are harmonic factors, and $\lVert \theta \rVert_2$ is the regularizer defined as $L_2$ distance.

$\mathcal{L}_{\theta_E, \theta_R}(G)$ is responsible to learn representations of both entities and relations from the knowledge graph $G$. This part will be introduced in detail in Section \ref{sec:kg}.

$\mathcal{L}_{\theta}(D)$ is responsible to learn representations of  entities and relations as well as words from the text corpus $T$. It is straightforward to learn word representations from text as discussed in Section \ref{sec:related}. On the contrary, since entities and relations are not explicitly shown in text, we have to identify entities and relations in text to support representation learning of entities and relations from text. The process is realized by entity-text alignment and relation-text alignment.

\textbf{Entity-Text Alignment.} Many entities are mentioned in text. Due to the complex polysemy of entity mentions (e.g., an entity name Washington in a sentence could be indicating either a person or a location), it is non-trivial to build entity-text alignment. The alignment can be built via entity linking techniques or anchor text information. In this paper, we simply use the anchor text annotated in articles to build the alignment between entities in $E$ and entity mentions in $V$. We will share the aligned entity representations to corresponding entity mentions.

\textbf{Relation-Text Alignment.} As mentioned in Section \ref{sec:related}, textual relations can be extracted from text. Hence, relation representation can also be learned from plain text. Inspired by the idea of distant supervision, for a relation $r \in R$, we collect all entity pairs $P_{r} = \{(h, t)\}$ connected by $r$ in KG. Afterwards, for each entity pair in $P_{r}$, we extract all sentences that contain the both entities from $D$, and regard them as the positive instances of the relation $r$. We can further apply deep neural networks to encode the semantic of these sentences into the corresponding relation representation. The process will be introduced in detail in Section \ref{sec:relation}.

In summary, the framework enables joint representation learning of both entities and relations by taking full advantages of both KG and text. The learned representations are expected to be more informative and robust, which will be verified in experiments.

\subsection{Representation Learning of KGs}
\label{sec:kg}
We select TransE \cite{bordes2013translating} to learn representations of entities and relations from KGs.

For each entity pair $(h, t)$ in a KG $G$, we define their latent relation embedding $\mathbf{r}_{ht}$ as a translation from $\mathbf{h}$ to $\mathbf{t}$, which can be formalized as:
\begin{equation}
\textbf{r}_{ht} = \textbf{t} - \textbf{h}.
\end{equation}
Meanwhile, each triple $(h, r, t) \in T$ has an explicit relation $r$ between $h$ and $t$. Hence, we can define the scoring function for each triple as follows:
\begin{equation}
f_r(h, t) = \lVert \textbf{r}_{ht} - \textbf{r} \rVert_2 = \lVert (\textbf{t} - \textbf{h}) - \textbf{r}  \rVert_2.
\end{equation}
This indicates that, for each triple $(h, r, t)$ in $T$, we expect $\textbf{h} + \textbf{r} \approx \textbf{t}$.

Based on the above scoring function, we can formalize the loss function over all triples in $T$ as follows:
\begin{equation}
\mathcal{L}(G) = \sum_{(h,r,t) \in T}\sum_{(h',r',t') \in T'} \big[ \gamma +f_r(h, t) - f_{r'}(h', t') \big]_{+}.
\label{eq:lg}
\end{equation}
Here $[x]_{+}$ indicates keeping the positive part of $x$ and $\gamma > 0$ is a margin. $T'$ is the set of incorrect triples:
\begin{equation}
T' = \{(h',r,t)\} \cup \{(h,r',t)\} \cup \{(h,r,t')\},
\end{equation}
which is constructed by replacing the entity and relation in each triple $(h, r, t) \in T$ with other entities $h', t' \in E$ and relations $r' \in R$.

\subsection{Representation Learning of Textual Relations}
\label{sec:relation}

Given a sentence containing two entities, the words in the sentence usually expose implicit features of the textual relation between the two entities. As shown in \cite{zeng2014relation}, the textual relations can be learned with deep neural networks and encoded in the low-dimensional semantic space.

We follow \cite{zeng2014relation} and apply convolutional neural networks (CNN) to model textual relations from text. CNN is an efficient neural model widely used in image processing, which has recently been verified to be also effective for many NLP tasks such as part-of-speech tagging, named entity recognition and semantic role labeling \cite{collobert2011natural}.

\begin{figure}[h]
\centering
\includegraphics[width=1\columnwidth]{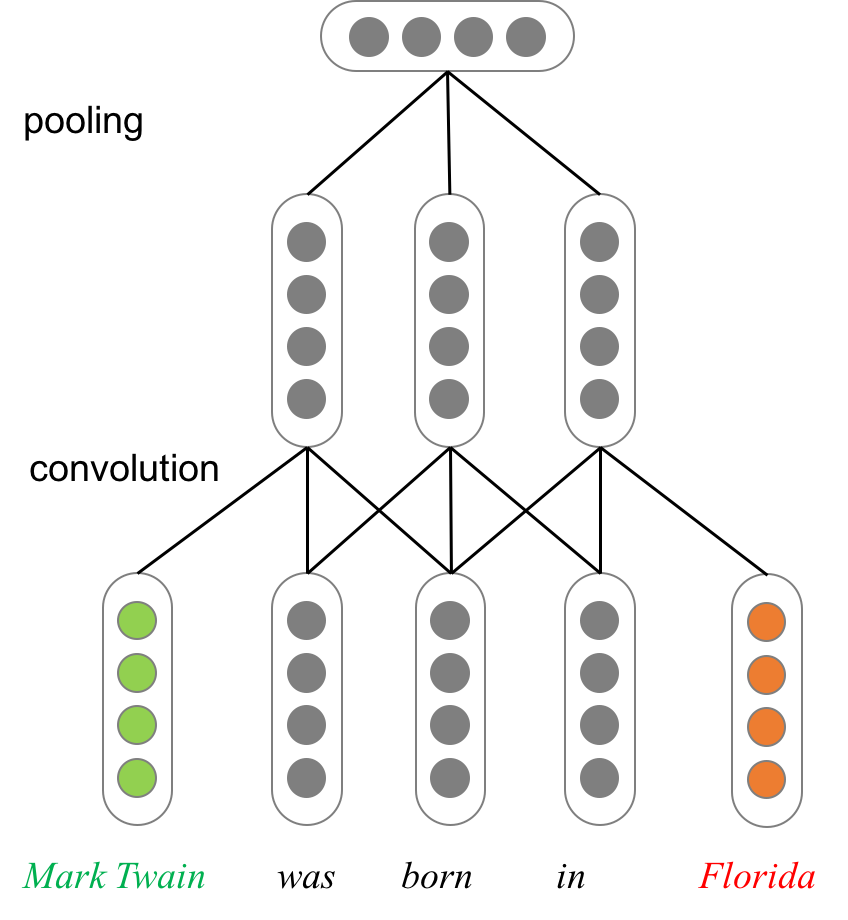}
\caption{Convolutional neural networks for representation learning of textual relations.}
\label{fig:cnn}
\end{figure}

\subsubsection{Overall Architecture}

Figure \ref{fig:cnn} depicts the overall architecture of CNN for modeling textual relations. For a sentence $s$ containing $(h, t)$ with a relation $r$, the architecture takes word embeddings $\mathbf{s} = \{\mathbf{x}_1, \ldots, \mathbf{x}_n \}$ of the sentence $s$ as input, and after passing through two layers within CNN, outputs the embedding of the textual relation $\mathbf{r}_{s}$. Our method will further learn to minimize the loss function between $\mathbf{r}$ and $\mathbf{r}_s$, which can be formalized as:
\begin{equation}
f_{r}(s) = \lVert \textbf{r}_{s} - \textbf{r} \rVert_2.
\label{eq:cnn_distance}
\end{equation}
Based on the scoring function, we can formalize the loss function over all sentences in $D$ as follows,
\begin{equation}
\mathcal{L}(D) = \sum_{s \in D}\sum_{r' \ne r} \big[ \gamma +f_r(s) - f_{r'}(s) \big]_{+},
\end{equation}
where the notations are identical to Eq. (\ref{eq:lg}).

CNN contains an input layer, a convolution layer and a pooling layer, which are introduced in detail as follows.

\subsubsection{Input Layer}
Given a sentence $s$ made up of $n$ words $s = \{ x_1, \ldots , x_n\}$, the input layer transforms the words of $s$ into corresponding word embeddings $\mathbf{s} = \{ \mathbf{x}_1, \ldots , \mathbf{x}_n \}$. For a word $x_i$ in the given sentence, its input embedding $\mathbf{x}_i$ is composed of two real-valued vectors: its textual word embedding $\mathbf{w}_i$ and its position embedding $\mathbf{p}_{i}$.

Textual word embeddings encode the semantics of the corresponding words, which are usually pre-trained from plain text via word representation learning, as introduced in Section \ref{sec:detail}.

Word position embeddings (WPE) is originally proposed in \cite{zeng2014relation}. WPE is a position feature indicating the relative distances of the given word to the marked entities in the sentence. As shown in Figure \ref{fig:cnn}, the relative distances of the word \emph{born} to the entities \emph{Mark Twain} and \emph{Florida} are $-2$ and $+2$ respectively. We map each distance to a vector of dimension $k_p$ in the continuous latent space. Given the word $x_i$ in the sentence $s$, the word position embedding is $\mathbf{p}_i = [\mathbf{p}^h_i, \mathbf{p}^t_i]$, where $\mathbf{p}^h_i$ and $\mathbf{p}^t_i$ are vectors of distances to the head entity and tail entity respectively.

We simply concatenate textual word embeddings and word position embeddings to build the input for CNN:
\begin{equation}
\mathbf{s} = \{[\mathbf{w}_1;\mathbf{p}_1],\ldots, [\mathbf{w}_n;\mathbf{p}_n]\}.
\end{equation}

\subsubsection{Convolution Layer}
By taking $\mathbf{s}$ as the input, the convolution layer will output $\mathbf{y}$. The generation process is formalized as follows.

We slide a window of size $m$ over the input word sequence. For each move, we can get an embedding $\mathbf{x}'_i$ as:
\begin{equation}
\mathbf{x}'_i = \big[ \mathbf{x}_{i - \frac{m-1}{2}}; \ldots ; \mathbf{x}_i; \ldots ;\mathbf{x}_{i + \frac{m-1}{2}} \big],
\end{equation}
which is obtained by concatenating $m$ vectors in $\mathbf{s}$ with $\mathbf{x}_i$ as center. For instance in Figure \ref{fig:cnn}, a window slides through the input vectors $\mathbf{s}$ and concatenates every three word embeddings. Afterwards, we transform $\mathbf{x}'_i$ into the hidden layer vector $\textbf{y}_i$
\begin{equation}
\mathbf{y}_i = \tanh(\mathbf{W}\mathbf{x}'_i + \mathbf{b}),
\end{equation}
where $\mathbf{W} \in \mathbb{R}^{k_c \times mk_w}$ is the convolution kernel, $\mathbf{b} \in \mathbb{R}^{k_c}$ is a bias vector, $k_c$ is the dimension of hidden layer vectors $\mathbf{y}_i$, $k_w$ is the dimension of input vectors $\textbf{x}_i$, and $m$ is the window size.

\subsubsection{Pooling Layer}
In the pooling layer, a max-pooling operation over the hidden layer vectors ${\mathbf{y}_1, \ldots , \mathbf{y}_n}$ is applied to get the final continuous vector as the textual relation embedding $\mathbf{r}_s$, which is formalized as follows:
\begin{equation}
\mathbf{r}_{s,j} = \max \{\mathbf{y}_{1,j}, \ldots, \mathbf{y}_{n, j} \},
\end{equation}
where $\mathbf{r}_{s,j}$ is the $j$-th value of the textual relation embedding $\mathbf{r}_s$, and $\mathbf{y}_{i,j}$ is the $j$-th value of the hidden layer vector $\mathbf{y}_i$. After the pooling operation, we can get the given sentence textual relation embedding to loss function Eq. (\ref{eq:cnn_distance}).

\subsection{Initialization and Implementation Details}
\label{sec:detail}
There are a large number of parameters to be optimized for joint learning. It is thus crucial to initialize these parameters appropriately. For those aligned entities and words, we initialize their embeddings via word representation learning. We follow \cite{mikolov2013linguistic} and use Skip-Gram to learn word representations from the given text corpus. For relations and other entities, we initialize their embeddings randomly.

Both the knowledge model TransE and textual relation model CNN are optimized simultaneously using stochastic gradient descent (SGD). The parameters of all models are trained using a batch training algorithm. Note that, the gradients of CNN parameters will be back-propagated to the input word embeddings so that the embeddings of both entities and words can also be learned from plain text via CNN.

\section{Experiments}

We conduct experiments on entity prediction and relation prediction and evaluate the performance of our methods with various baselines.

\subsection{Experiment Settings}

\subsubsection{Datasets}


\textbf{Knowledge Graph.} We select Freebase \cite{bollacker2008freebase} as the knowledge graph for joint learning. Freebase is a widely-used large-scale world knowledge graph. In this paper, we adopt a datasets extracted Freebase, FB15K, in our experiments. The dataset has been used in many studies on knowledge representation learning \cite{bordes2013translating,bordes2014semantic,lin2015learning}. We list the statistics of FB15K in Table \ref{statistics-of-FB15K}, including the amount of entities, relations and triples.

\begin{table}[htb]
\centering
\scriptsize
\caption{The statistics of FB15K}
\begin{tabular}{|c|c|c|c|c|c|}
\hline
Dataset & \textbf{Relation} & \textbf{Entity} & \textbf{Train} & \textbf{Valid}  & \textbf{Test}   \\ \hline
FB15K   & 1,345           & 14,951            & 483,142        & 50,000 & 59,071 \\ \hline
\end{tabular}
\label{statistics-of-FB15K}
\end{table}

\textbf{Text Corpus.} We select sentences from the New York Times articles to align with FB15K for joint learning. To ensure alignment accuracy, we only consider those sentences with anchor text linking to the entities in FB15K. We extract $876,227$ sentences containing both head and tail entities in FB15K triples, and annotate with the corresponding relations in triples. The sentences are labeled with $29,252$ FB15K triples, including $629$ relations and $5244$ entities. We name the corpus as NYT.

\subsubsection{Evaluation Tasks}

In experiments we evaluate the joint learning model and other baselines with three tasks:

(1) \textbf{Entity Prediction.} The task aims at predicting missing entities in a triple according to the embeddings of another entity and relation. 

(2) \textbf{Relation Prediction.} The task aims at predicting missing relations in a triple according to the embeddings of head and tail entities. 

(3) \textbf{Relation Classification from Text.} We are also interested in extracting relational facts between novel entities not included in knowledge graphs. Hence, we conduct relation classification from text, without taking advantages of entity embeddings learned with knowledge graph structure.

\subsubsection{Parameter Settings}

In our joint model, we select the learning rate $\alpha_k$ on the knowledge side among $\{0.1, 0.01, 0.001\}$, and learning rate $\alpha_t$ on the text side among $\{0.01, 0.025, 0.05\}$. The harmonic factor $\lambda = 1$ and the margin $\gamma = 1$. We select the harmonic factor $\tau$ among $\{0.001, 0.0001, 0.00001\}$ to balance the learning ratio between knowledge and text. The dimension of embeddings $k$ is selected among $\{50, 100, 150\}$. The optimal configurations are $\alpha_k = 0.001, \alpha_t = 0.025, \tau = 0.0001$, $k = 150$. During the learning process, we traverse the text corpus for $10$ rounds as well as triples in the knowledge graph for $3000$ rounds.

\begin{table*}[t]
\centering
\scriptsize
\begin{tabular}{|c|c|c|c|c|c|c|c|c|c|c|c|c|}
\hline
Metric            & \multicolumn{4}{c|}{Predicting Head} & \multicolumn{4}{c|}{Predicting Tail} & \multicolumn{2}{c|}{Overall} \\ \hline
                  & 1-to-1     & 1-to-N    & N-to-1    & N-to-N    & 1-to-1     & 1-to-N    & N-to-1    & N-to-N  & Triple Avg. & Relation Avg. \\ \hline
TransE            & 43.7       & 65.7      & 18.2      & 47.2      & 43.7       & 19.7      & 66.7      & 50.0    & 47.1 & - \\ \hline
TransH (unif)     & 66.7       & 81.7      & 30.2      & 57.4      & 63.7       & 30.1      & 83.2      & 60.8    & 58.5 & - \\ \hline
TransH (bern)     & 66.8       & 87.6      & 28.7      & 64.5      & 65.5       & 39.8      & 83.3      & 67.2    & 64.4 & - \\ \hline
TransR (unif)     & 76.9       & 77.9      & 38.1      & 66.9      & 76.2       & 38.4      & 76.2      & 69.1    & 65.5 & - \\ \hline
TransR (bern)     & 78.8       & 89.2      & 34.1      & 69.2      & 79.2       & 37.4      &\textbf{90.4}& 72.1  & 68.7 & - \\ \hline
TransE (Our)      & 66.5       & 88.8      & 39.8      & 79.0      & 66.4       & 51.9      & 85.6      & 81.5    & 76.6 & 66.2 \\ \hline
Joint             & \textbf{82.7}& \textbf{89.1}& \textbf{45.0}& \textbf{80.7}& \textbf{81.7}& \textbf{57.7}& 87.4&\textbf{82.8} & \textbf{78.7} & \textbf{79.1} \\ \hline
\end{tabular}
\caption{Evaluation results on entity prediction of head and tail entities (\%).}
\label{t:entity}
\end{table*}

\subsection{Results of Entity Prediction}

Entity prediction has also been used for evaluation in \cite{bordes2013translating,wang2014transh,lin2015learning}. More specifically, we need to predict the tail entity when given a triple $(h, r, ?)$ or predict the head entity when given a triple $(?, r ,t)$. In this task, for each missing entity, the system is asked to rank all candidate entities from the knowledge graph instead of only giving one best result. For each test triple $(h, r, t)$, we replace head and tail entities with all entities in FB15K ranked in descending order of similarity scores calculated by $\lVert \textbf{h} + \textbf{r} - \textbf{t} \rVert_2$. The relational fact $(h, r, t)$ is expected to have smaller score than any other corrupted triples.

We follow \cite{bordes2013translating,wang2014transh,lin2015learning} and use the proportion of correct entities in Top-10 ranked entities (Hits@10) as the evaluation metric. As mentioned in \cite{bordes2013translating}, a corrupted triple may also exist in knowledge graphs, which should not be considered as incorrect. Hence, before ranking, we filter out those corrupted triples that have appeared in FB15K.

The relations in knowledge graphs can be divided into four classes: 1-to-1, 1-to-N, N-to-1 and N-to-N relations, where a ``1-to-N'' relation indicates a head entity may correspond to multiple tail entities in knowledge graphs, and so on. For example, the relation (\emph{Country}, \texttt{PresidentOf}, \emph{Person}) is a typical ``1-to-N'' relation, because there used to be many presidents for a country in history. We report the average Hits@10 scores when predicting missing head entities and tail entities with respect to different classes of relations. We also report the overall performance by averaging the Hits@10 scores over triples and over relations.

Since the evaluation setting is identical, we simply report the results of TransE, TransH and TransR from \cite{bordes2013translating,wang2014transh,lin2015learning}, where ``unif'' and ``bern'' are two settings to sample negative instances for learning. We also report the results of TransE we implement for joint learning.
 
The evaluation results on entity prediction is shown in Table \ref{t:entity}. From Table \ref{t:entity} we observe that:

(1) The joint model almost achieves improvements under four classes of relations when predicting head and tail entities. This indicates the performance of joint learning is consistent and robust. Note that, our TransE version, which is implemented by ourselves, outperforms previous TransE, TransH and TranR, by simply increase the embedding dimension to $k=150$, which suggests the effectiveness of TransE.

(2) The improvements on ``1-to-1'', ``1-to-N'' and ``N-to-1'' relations are much more significant as compared to those on ``N-to-N''. This indicates that our joint model is more effective to embed textual relations for those deterministic relations.

(3) Our joint model achieves improvement of more than $13\%$ than TransE when averaging over relations. This indicates that, our joint model can take advantages of plain texts and greatly improve representation power in relation-level.

(4) In FB15K, the relation numbers in different relation classes are comparable, but more than $80\%$ triples are instances of ``N-to-N'' relations. Since the improvement of the joint model on ``N-to-N'' relations is not as remarkable as on other relation classes, hence the overall superiority of our joint model seems not so notable when averaging over triples as compared to averaging over relations.

\subsection{Results of Relation Prediction}
The task aims to predict the missing relation between two entities based on their embeddings. More specifically, we need to predict the relation when given a triple $(h, ?, t)$. In this task, for each missing relation, the system is asked to find one best result, according to similarity scores calculated by $\lVert \textbf{h} + \textbf{r} - \textbf{t} \rVert_2$.
Because the number of relations is much smaller, compared with the number of entities, we use the accuracy of Top-1 ranked relations as the evaluation metric. Since some entities may have more than one relation between them, we also filter out those triples with corrupted relations appeared in knowledge graphs. We report the overall evaluation results as well as those in different relation classes.

\begin{table}[htb]
\centering
\scriptsize
\begin{tabular}{|c|c|c|c|c|c|}
\hline
Tasks             & \multicolumn{5}{c|}{Relation Prediction}                      \\ \hline
Category & 1-to-1     & 1-to-N     & N-to-1     & N-to-N     & All           \\ \hline
TransE(Our)       & 24.1       & 83.0       & 80.4       & 92.5       & 87.2          \\ \hline
Joint             & \textbf{40.9} & \textbf{89.4} & \textbf{87.1} & \textbf{94.6} & \textbf{91.6} \\ \hline
\end{tabular}
\caption{Evaluation results on relation prediction (\%).}
\label{t:relation}
\end{table}

The evaluation results are shown in Table \ref{t:relation}. From Table \ref{t:relation} we observe that, our joint model outperforms TransE consistently in different classes of relations and in all. The joint model also achieves more significant improvements on ``1-to-1'', ``1-to-N'' and ``N-to-1'' relations. The observations are compatible with those on entity prediction.

\subsection{Results of Relation Classification from Text}

The task aims to extract relational facts from plain text. The task has been widely studied, also named as \emph{relation extraction} from text. Most models \cite{mintz2009distant,riedel2010modeling,hoffmann2011knowledge,surdeanu2012multi} take knowledge graphs as distant supervision to automatically annotate sentences in text corpora as training instances, and then extract textual features to build relation classifiers. As compared to relation prediction with embeddings, the task only uses plain text to identify relational facts, and thus is capable for novel entities not necessarily having appeared in knowledge graphs. Since there is much noise in plain text and distant supervision, it makes the task not easy. With this task, we want to investigate the effectiveness of our joint model for learning CNN models.

We follow \cite{weston2013connecting} to conduct evaluation. The evaluation construct candidate triples combined by entity pairs in testing set and various relations, ask systems to rank these triples according to the corresponding sentences of entity pairs, and by regarding the triples in knowledge graphs as correct and others as incorrect, evaluate systems with precision-recall curves. Note that, the evaluation task does not consider knowledge embeddings for ranking.

\begin{figure}[]
\centering
\includegraphics[width=1\columnwidth]{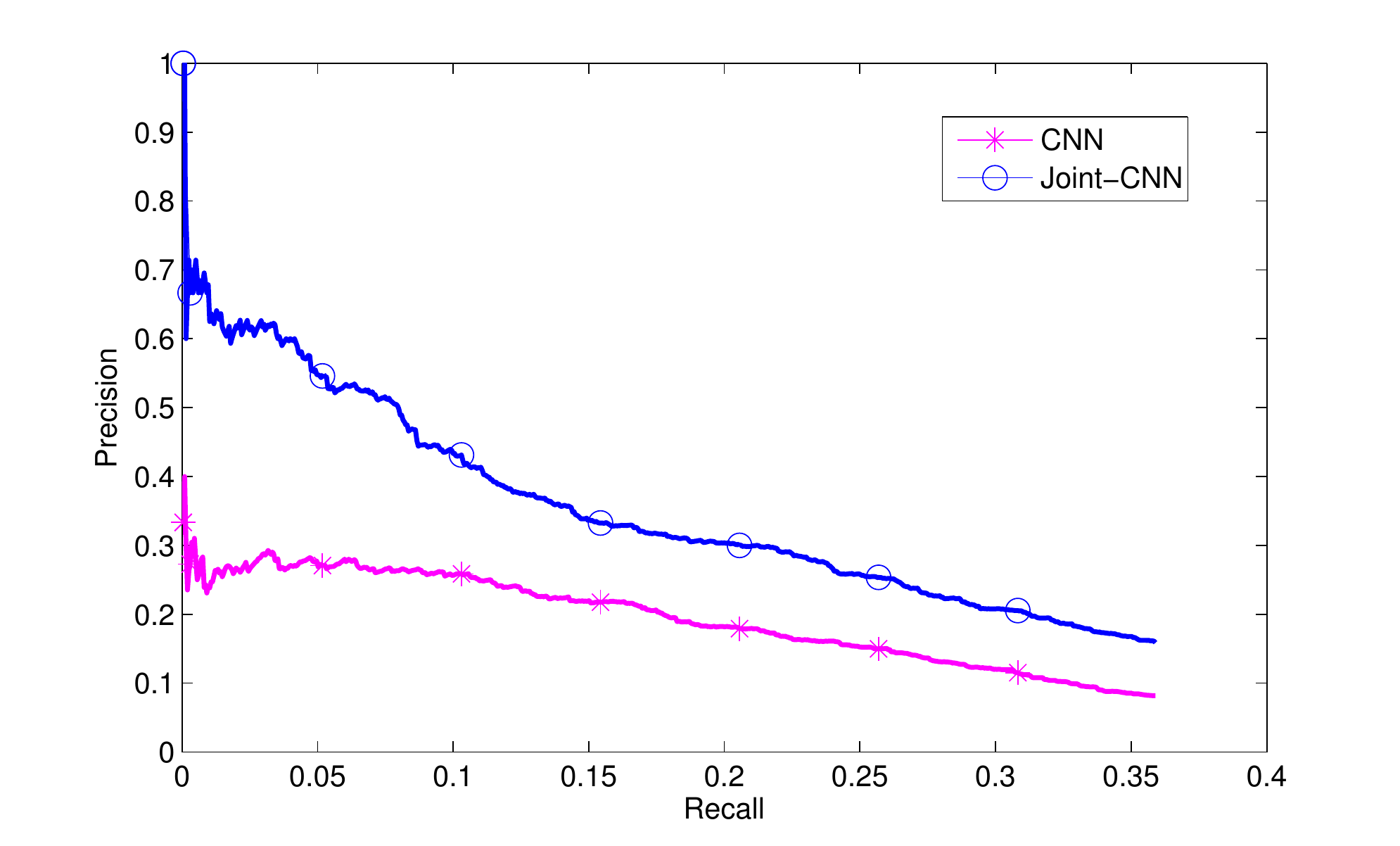}
\caption{Evaluation results on relation classification from text.}
\label{fig:relation_big}
\end{figure}

The evaluation results on NYT test set are shown in Figure \ref{fig:relation_big}, where Joint-CNN indicates the CNN model learned jointly in our model, and CNN indicates the conventional CNN model learned individually from plain text. We find that, the sentence counts of different relation types vary much, and may also influence the performance of relation classification. About ninety-five percent sentences belong to the most frequent 100 relations in our dataset. In order to alleviate the influence of sentence counts, we select Top-$100$ relations and evaluate the classification performance among them. From Figure \ref{fig:relation_big} we observe that: Joint-CNN outperforms CNN significantly over all the range. This indicates that, the joint learning model can also result in a more effective CNN model for relation classification from text. This will greatly benefit the relation extraction task, especially for those novel entities.

\section{Conclusion and Future Work}

In this paper, we propose a model for joint learning of text and knowledge representations. Our joint model embeds entities, relations and words in the same continuous latent space. More specifically, we adopt deep neural networks CNN to encode textual relations for joint learning of relation embeddings. In experiments, we evaluate our joint model on three tasks including entity prediction, relation prediction with embeddings, and relation prediction from text. Experiment results show that our joint model can effectively perform representation learning from both knowledge graphs and plain text, and obtain more discriminative entity and relation embeddings for prediction. In future, we will explore the following research directions: 

(1) Distant supervision may introduce many noisy sentences with incorrect relation annotations. We will explore techniques such as multi-instance learning to reduce these noises and improve the effectiveness of joint learning. We will also explore the effectiveness of more deep neural networks like recurrent neural networks, long short-term memory other than CNN for joint learning. 

(2) Our joint model is also capable to incorporate other knowledge representation models instead of TransE, such as TransH and TransR. In future we will explore their capability in our joint model. 

(3) We will also take more rich information in our joint model, such as relation paths in knowledge graphs, and the textual relations represented by more than one sentence in a paragraph or document. These information can also be used to incorporate into knowldege graphs.

These future work will further improve performance over knowledge and text representation, this may let the joint model make better use of knowledge and text.

\bibliography{emnlp2016}

\begin{thebibliography}{}

\bibitem[\protect\citename{Auer \bgroup et al.\egroup }2007]{auer2007dbpedia}
S{\"o}ren Auer, Christian Bizer, Georgi Kobilarov, Jens Lehmann, Richard
  Cyganiak, and Zachary Ives.
\newblock 2007.
\newblock {\em Dbpedia: A nucleus for a web of open data}.
\newblock Springer.

\bibitem[\protect\citename{Bollacker \bgroup et al.\egroup
  }2008]{bollacker2008freebase}
Kurt Bollacker, Colin Evans, Praveen Paritosh, Tim Sturge, and Jamie Taylor.
\newblock 2008.
\newblock Freebase: a collaboratively created graph database for structuring
  human knowledge.
\newblock In {\em Proceedings of KDD}, pages 1247--1250.

\bibitem[\protect\citename{Bordes \bgroup et al.\egroup
  }2013]{bordes2013translating}
Antoine Bordes, Nicolas Usunier, Alberto Garcia-Duran, Jason Weston, and Oksana
  Yakhnenko.
\newblock 2013.
\newblock Translating embeddings for modeling multi-relational data.
\newblock In {\em Proceedings of NIPS}, pages 2787--2795.

\bibitem[\protect\citename{Bordes \bgroup et al.\egroup
  }2014]{bordes2014semantic}
Antoine Bordes, Xavier Glorot, Jason Weston, and Yoshua Bengio.
\newblock 2014.
\newblock A semantic matching energy function for learning with
  multi-relational data.
\newblock {\em Machine Learning}, 94(2):233--259.

\bibitem[\protect\citename{Collobert \bgroup et al.\egroup
  }2011]{collobert2011natural}
Ronan Collobert, Jason Weston, L{\'e}on Bottou, Michael Karlen, Koray
  Kavukcuoglu, and Pavel Kuksa.
\newblock 2011.
\newblock Natural language processing (almost) from scratch.
\newblock {\em JMLR}, 12:2493--2537.

\bibitem[\protect\citename{dos Santos \bgroup et al.\egroup
  }2015]{dos2015classifying}
C{\i}cero~Nogueira dos Santos, Bing Xiang, and Bowen Zhou.
\newblock 2015.
\newblock Classifying relations by ranking with convolutional neural networks.
\newblock In {\em Proceedings of ACL-IJCNLP}, volume~1, pages 626--634.

\bibitem[\protect\citename{Erhan \bgroup et al.\egroup }2010]{erhan2010does}
Dumitru Erhan, Yoshua Bengio, Aaron Courville, Pierre-Antoine Manzagol, Pascal
  Vincent, and Samy Bengio.
\newblock 2010.
\newblock Why does unsupervised pre-training help deep learning?
\newblock {\em JMLR}, 11:625--660.

\bibitem[\protect\citename{Hoffmann \bgroup et al.\egroup
  }2011]{hoffmann2011knowledge}
Raphael Hoffmann, Congle Zhang, Xiao Ling, Luke Zettlemoyer, and Daniel~S Weld.
\newblock 2011.
\newblock Knowledge-based weak supervision for information extraction of
  overlapping relations.
\newblock In {\em Proceedings of ACL-HLT}, pages 541--550.

\bibitem[\protect\citename{Lao and Cohen}2010]{lao2010relational}
Ni~Lao and William~W Cohen.
\newblock 2010.
\newblock Relational retrieval using a combination of path-constrained random
  walks.
\newblock {\em Machine learning}, 81(1):53--67.

\bibitem[\protect\citename{Lao \bgroup et al.\egroup }2011]{lao2011random}
Ni~Lao, Tom Mitchell, and William~W Cohen.
\newblock 2011.
\newblock Random walk inference and learning in a large scale knowledge base.
\newblock In {\em Proceedings of EMNLP}, pages 529--539.

\bibitem[\protect\citename{Lin \bgroup et al.\egroup }2015]{lin2015learning}
Yankai Lin, Zhiyuan Liu, Maosong Sun, Yang Liu, and Xuan Zhu.
\newblock 2015.
\newblock Learning entity and relation embeddings for knowledge graph
  completion.
\newblock In {\em Proceedings of AAAI}.

\bibitem[\protect\citename{Mikolov \bgroup et al.\egroup
  }2013a]{mikolov2013efficient}
Tomas Mikolov, Kai Chen, Greg Corrado, and Jeffrey Dean.
\newblock 2013a.
\newblock Efficient estimation of word representations in vector space.
\newblock {\em Proceedings of ICLR}.

\bibitem[\protect\citename{Mikolov \bgroup et al.\egroup
  }2013b]{mikolov2013distributed}
Tomas Mikolov, Ilya Sutskever, Kai Chen, Greg~S Corrado, and Jeff Dean.
\newblock 2013b.
\newblock Distributed representations of words and phrases and their
  compositionality.
\newblock In {\em Proceedings of NIPS}, pages 3111--3119.

\bibitem[\protect\citename{Mikolov \bgroup et al.\egroup
  }2013c]{mikolov2013linguistic}
Tomas Mikolov, Wen-tau Yih, and Geoffrey Zweig.
\newblock 2013c.
\newblock Linguistic regularities in continuous space word representations.
\newblock In {\em Proceedings of HLT-NAACL}, pages 746--751.

\bibitem[\protect\citename{Miller}1995]{miller1995wordnet}
George~A Miller.
\newblock 1995.
\newblock Wordnet: a lexical database for english.
\newblock {\em Communications of the ACM}, 38(11):39--41.

\bibitem[\protect\citename{Mintz \bgroup et al.\egroup }2009]{mintz2009distant}
Mike Mintz, Steven Bills, Rion Snow, and Dan Jurafsky.
\newblock 2009.
\newblock Distant supervision for relation extraction without labeled data.
\newblock In {\em Proceedings of ACL-IJCNLP}, pages 1003--1011.

\bibitem[\protect\citename{Riedel \bgroup et al.\egroup
  }2010]{riedel2010modeling}
Sebastian Riedel, Limin Yao, and Andrew McCallum.
\newblock 2010.
\newblock Modeling relations and their mentions without labeled text.
\newblock In {\em Machine Learning and Knowledge Discovery in Databases}, pages
  148--163.

\bibitem[\protect\citename{Suchanek \bgroup et al.\egroup
  }2007]{suchanek2007yago}
Fabian~M Suchanek, Gjergji Kasneci, and Gerhard Weikum.
\newblock 2007.
\newblock Yago: a core of semantic knowledge.
\newblock In {\em Proceedings of WWW}, pages 697--706. ACM.

\bibitem[\protect\citename{Surdeanu \bgroup et al.\egroup
  }2012]{surdeanu2012multi}
Mihai Surdeanu, Julie Tibshirani, Ramesh Nallapati, and Christopher~D Manning.
\newblock 2012.
\newblock Multi-instance multi-label learning for relation extraction.
\newblock In {\em Proceedings of EMNLP}, pages 455--465.

\bibitem[\protect\citename{Toutanova \bgroup et al.\egroup
  }2015]{toutanova2015representing}
Kristina Toutanova, Danqi Chen, Patrick Pantel, Pallavi Choudhury, and Michael
  Gamon.
\newblock 2015.
\newblock Representing text for joint embedding of text and knowledge bases.
\newblock In {\em Proceedings of EMNLP}.

\bibitem[\protect\citename{Wang \bgroup et al.\egroup
  }2014a]{wang2014knowledge}
Zhen Wang, Jianwen Zhang, Jianlin Feng, and Zheng Chen.
\newblock 2014a.
\newblock Knowledge graph and text jointly embedding.
\newblock In {\em Proceedings of EMNLP}, pages 1591--1601.

\bibitem[\protect\citename{Wang \bgroup et al.\egroup }2014b]{wang2014transh}
Zhen Wang, Jianwen Zhang, Jianlin Feng, and Zheng Chen.
\newblock 2014b.
\newblock Knowledge graph embedding by translating on hyperplanes.
\newblock In {\em Proceedings of AAAI}, pages 1112--1119.

\bibitem[\protect\citename{Weston \bgroup et al.\egroup
  }2013]{weston2013connecting}
Jason Weston, Antoine Bordes, Oksana Yakhnenko, and Nicolas Usunier.
\newblock 2013.
\newblock Connecting language and knowledge bases with embedding models for
  relation extraction.

\bibitem[\protect\citename{Xu \bgroup et al.\egroup }2015]{xu2015classifying}
Yan Xu, Lili Mou, Ge~Li, Yunchuan Chen, Hao Peng, and Zhi Jin.
\newblock 2015.
\newblock Classifying relations via long short term memory networks along
  shortest dependency paths.
\newblock In {\em Proceedings of EMNLP}.

\bibitem[\protect\citename{Zeng \bgroup et al.\egroup }2014]{zeng2014relation}
Daojian Zeng, Kang Liu, Siwei Lai, Guangyou Zhou, and Jun Zhao.
\newblock 2014.
\newblock Relation classification via convolutional deep neural network.
\newblock In {\em Proceedings of COLING}, pages 2335--2344.

\bibitem[\protect\citename{Zhang and Wang}2015]{zhang2015relation}
Dongxu Zhang and Dong Wang.
\newblock 2015.
\newblock Relation classification via recurrent neural network.
\newblock {\em arXiv preprint arXiv:1508.01006}.

\end{thebibliography}
\bibliographystyle{emnlp2016}

\end{document}